\documentclass{article} 
\usepackage[preprint]{colm2025_conference}

\usepackage{microtype}
\usepackage{hyperref}
\usepackage{url}
\usepackage{booktabs}
\usepackage{graphicx}
\usepackage{multirow}
\usepackage{lineno}
\usepackage{enumitem}
\usepackage{wrapfig}
\usepackage{amsmath}

\makeatletter
\g@addto@macro\bfseries{\boldmath}
\makeatother

\definecolor{darkblue}{rgb}{0, 0, 0.5}
\hypersetup{colorlinks=true, citecolor=darkblue, linkcolor=darkblue, urlcolor=darkblue}

\usepackage[breakable]{tcolorbox}
\usepackage{listings}
\newcommand{\gpt}[0]{GPT-4o}
\newcommand{\llama}[0]{Llama3.1-8B-Instruct}
\newcommand{\llamabase}[0]{Llama3.1-8B-Base}
\newcommand{\plus}[1]{{\footnotesize\color{blue}(+#1\%)}}

\usepackage[capitalise]{cleveref}

\crefname{section}{Sec.}{Secs.}
\Crefname{section}{Section}{Sections}
\Crefname{table}{Table}{Tables}
\crefname{table}{Table}{Tables}

\usepackage{xspace}

\title{Executable Functional Abstractions: Inferring Generative \\ Programs for Advanced Math Problems}

\author{Zaid Khan,~~Elias Stengel-Eskin,~~Archiki Prasad,~~Jaemin Cho,~~Mohit Bansal \vspace{5pt}\\
University of North Carolina at Chapel Hill\\
\texttt{\{zaidkhan, esteng, archiki, jmincho, mbansal\}@cs.unc.edu}
}

\newcommand{\method}{\texttt{EFAGen}\xspace}

\newcommand{\efa}{\texttt{EFA}\xspace}
\newcommand{\efas}{\texttt{EFAs}\xspace}

\newtcolorbox[auto counter,number within=section,crefname={box}{boxes}]{pabox}[2][]{%
title=Box 1 $\mid$,
colback=gray!10, colframe=black, sharp corners, breakable,
subtitle style={boxrule=0.4pt,colback=cyan!50!red!25!white},title=Box ~\thetcbcounter $\mid$ #2, label={#1}}

\begin{document}

\ifcolmsubmission
\linenumbers
\fi

\maketitle

\begin{abstract}
Scientists often infer abstract procedures from specific instances of problems and use the abstractions to generate new, related instances.
For example, programs encoding the formal rules and properties
of a system have been useful in fields ranging from reinforcement learning (procedural environments) to physics (simulation engines).
These programs 
can be seen as functions which execute to different outputs based on their parameterizations (e.g., gridworld configuration or initial physical conditions).
We introduce the term \efa{} (\textbf{E}xecutable \textbf{F}unctional \textbf{A}bstraction) to denote such programs for math problems. 
\efa{}-like constructs have been shown to be useful for mathematical reasoning as problem generators for stress-testing models. 
However, prior work has been limited to automatically constructing  abstractions for grade-school math (whose simple rules are easy to encode in programs), while generating \efa{}s for advanced math has thus far required human engineering.
We explore the automatic construction of \efa{}s for advanced mathematics problems by developing \method{}, which operationalizes the task of automatically inferring an \efa{} for a given seed problem and solution as a program synthesis task.
We first formalize the properties of any valid \efa{}~as executable unit tests.
Using execution feedback from the unit tests, we search over candidate programs sampled from a large language model (LLM) to find \efa{}~programs that are faithful to the generalized problem and solution class underlying the seed problem.
We then apply the tests as a reward signal, training LLMs to become better writers of \efa{}s.
We show that \efa{}s inferred by \method{}~are faithful to the seed problems, produce learnable problem variations, and that 
\method{} can infer \efa{}s across diverse sources of competition-level math problems.
Finally, we show uses of model-written \efa{}s, such as finding problem variations that are harder or easier for a learner to solve, as well as data generation.
\footnote{Code, models, and data at \href{https://zaidkhan.me/EFAGen}{zaidkhan.me/EFAGen}}

\end{abstract}

\vspace{-0.25em}
\section{Introduction}
\label{sec:intro}
\vspace{-0.25em}
In many fields, experts abstract specific instances into general procedures that can generate a wide range of related cases.
For example, physicists distill observations of falling objects into equations of motion capable of predicting trajectories under varying initial conditions~\citep{sep-newton-principia}.
This ability is not limited to certain domain experts: in fact, the ability to infer underlying compositional structures from surface forms is a core component of human language and intelligence \citep{chomsky1957syntactic, montague1970universal, partee2008compositionality, lake2017building}. 
The outcome of this process of abstraction is often a data-generating program whose execution is controlled by parameters, such as a gridworld generator that produces different world layouts given different configuration files. 
In fields such as reinforcement learning,  notable instances of data generating programs such as Holodeck \citep{holodeck} and BabyAI \citep{babyai} have become important parts of the research ecosystem for their capability to endlessly generate well-formed randomized task instances.

\begin{figure}[t]
    \centering
    \includegraphics[width=1.0\linewidth]{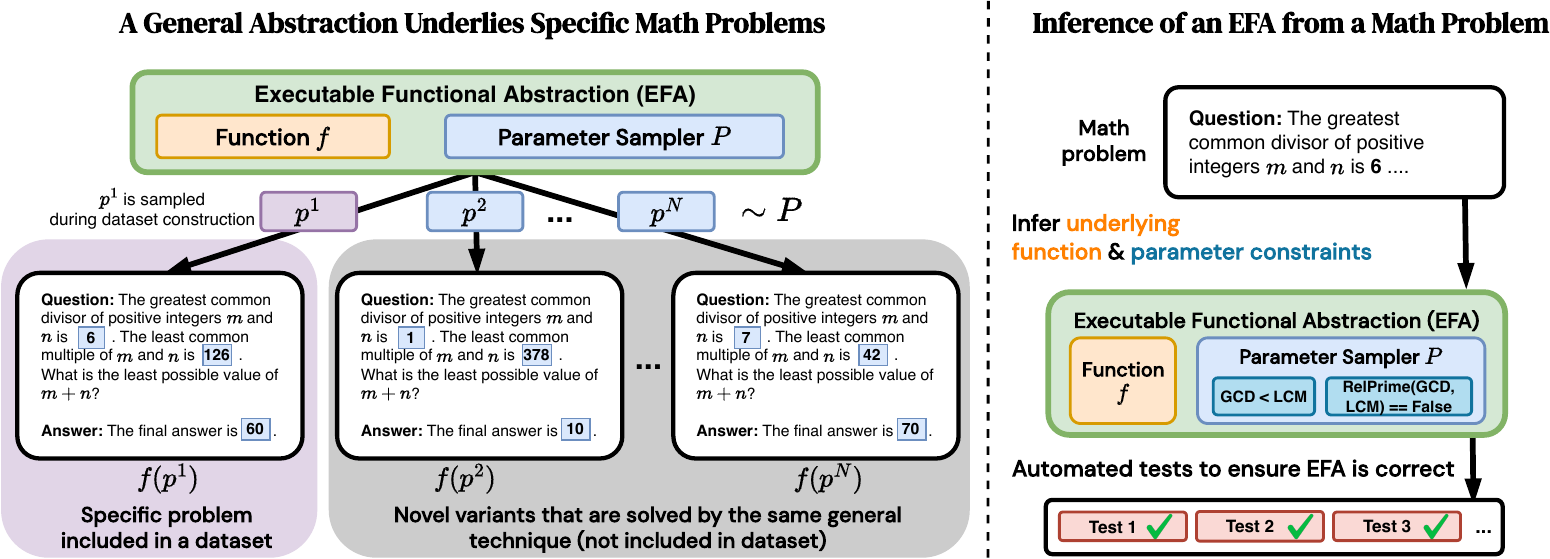}
    \vspace{-1em}
    \caption{
    \textbf{Left}: 
    The generative process underlying computational math problems,
    where the different instances share the same underlying problem-solving logic (function) but differ in parameter values.
    We introduce \textbf{executable functional abstractions (\efa{}s)}
    to model this latent structure.
    \textbf{Right}: 
    we study the task of inferring \efa{}s; i.e., recovering the underlying problem-solving function and parameters from math problems expressed in natural language.
    }
    \label{fig:teaser}
    \vspace{-1em}
\end{figure}

We introduce \textbf{E}xecutable \textbf{F}unctional \textbf{A}bstraction (\efa{}), a programmatic abstraction that encapsulates the logic of a math problem in a parameterized form and enables the automated sampling of problems variants.
Although similar abstractions have been used in other domains, automatic construction of \efa{}s for generating fresh, diverse math problems remains largely unexplored.
The property enabling the construction of \efa{}s for mathematics is that many math problems are a surface form of a more abstract deep structure. 
For example, consider the problem in \cref{fig:teaser} (left), which asks for positive integers $m$ and $n$ with a greatest common divisor (GCD) of 6 and a least common multiple (LCM) of 126, seeking the minimum value of $m+n$, which we denote as \texttt{LcmGcdMinSum(gcd=6, lcm=126)}. 
This specific problem is a special case of a more general problem \texttt{LcmGcdMinSum(gcd=$g$, lcm=$l$)} where $l,g \in \mathbb{N}$ can be any natural numbers.
Inferring an \efa{} requires transforming the \texttt{LcmGcdMinSum(gcd=6, lcm=126)} problem about a specific pair of numbers into a program that generates valid LcmGcdMinSum problems with varying parameters while implementing a general solution procedure that solves any specific instances of the general problem, such as \texttt{LcmGcdMinSum(gcd=7, lcm=42)}.
In this paper, we explore the automatic creation of \efa{}s for higher-level math problems.
This leads us to our central research question: 
\begin{quote}
\textit{   How can we automatically transform static math problems into their corresponding executable functional abstractions (\efa{}s)? }
\end{quote}

\begin{figure}
    \centering
    \includegraphics[width=0.9\linewidth]{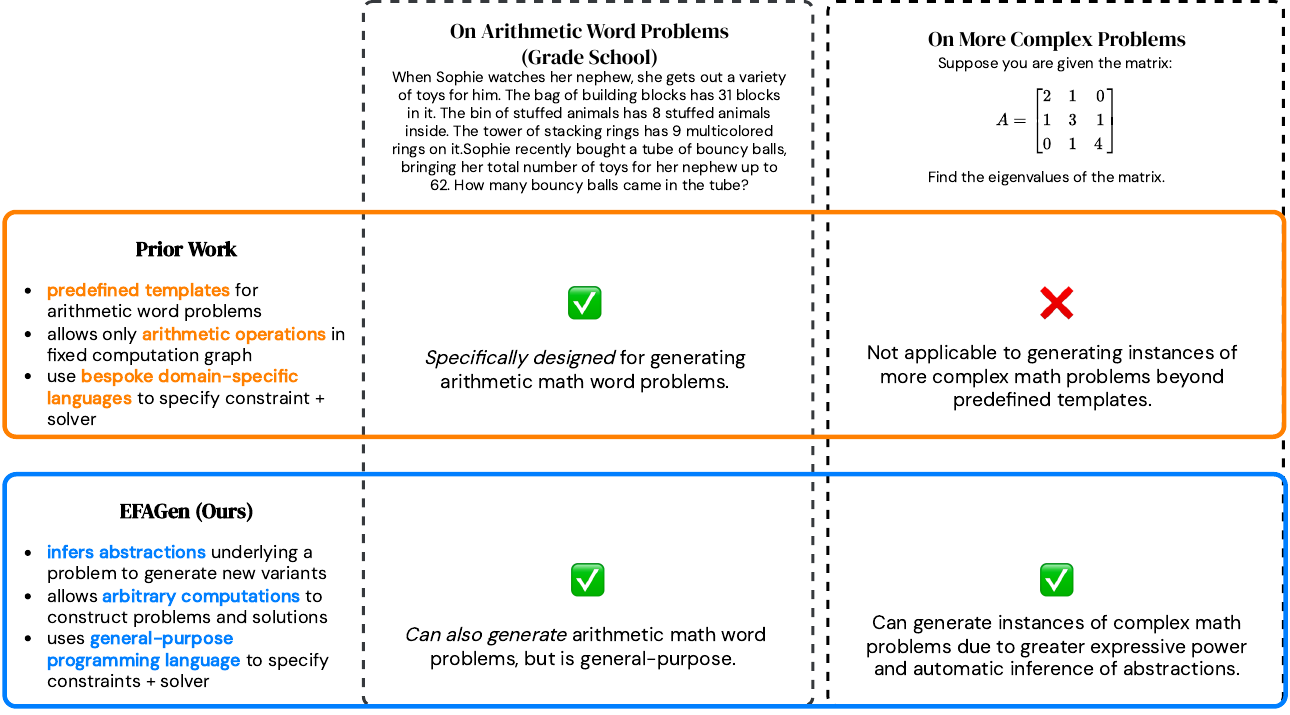}
    \textcolor{black}{\caption{\textbf{\method generalizes prior work on constructing arithmetic word problems to automatically constructing more complex, higher-level math problems}. Given a math problem and solution, \method~infers an underlying abstraction whose construction and general solution may involve arbitrary computations beyond fixed sequences of arithmetic operations.
    For example, the abstraction underlying the eigenvalue problem on the right is that of a tridiagonal $3 \times 3$ matrix. The general solution requires a symbolic computation composed with a numerial root-finding procedure. Details of inferred \efa{}~code in \cref{fig:expanded-efas-for-arithmetic-and-tridiagonal}.}
    \label{fig:efa-vs-template-based}}
    \vspace{-7mm}
\end{figure}

The task of automatically transforming static math problems into an \efa~is nontrivial. Recent work has made progress with grade-school level math problems~\citep{zhang2024darg,mirzadeh2025gsmsymbolic} by taking advantage of the simple computational graphs of their solutions. Higher-level problems with more complex computational graphs have thus far required human involvement to lift problems into functional forms~\citep{shah2024aiassisted,srivastava2024functional}.
An automated approach for mathematical problems more complex than grade-school arithmetic has not been developed. 
Such automatic construction of \efa{}s requires simultaneously solving multiple subproblems: identifying which numerical values should be parameterized, discovering the constraints between these parameters to maintain problem validity, abstracting the solution procedure to handle all valid parameterizations, and ensuring mathematical correctness across the entire parameter space. 
For example, in \cref{fig:teaser}, $m$ and $n$ are not parameters of the problem despite already being abstract variables, as they are dependent on the values of the \texttt{gcd} and \texttt{lcm} given. 
Nor can the \texttt{gcd} or \texttt{lcm} values be allowed to vary arbitrarily. 
Some parameterizations of the \texttt{gcd} and \texttt{lcm} may yield trivial problems (if the \texttt{gcd} is 1 and the \texttt{lcm} is a prime), while other parameterizations are simply invalid (such as \texttt{gcd} $>$ \texttt{lcm} or \texttt{gcd} and \texttt{lcm} being relatively prime).

We operationalize the task of inferring \efas as a program synthesis task using large language models (LLMs). Our method, \method, conditions an LLM on a static seed math problem and its step-by-step solution to generate candidate programs implementing an \efa~for the seed math problem. To generate a correct \efa{}, the program synthesizer must identify which numerical values in the static problem should be treated as parameters, determine appropriate sampling distributions for these parameters, and encode the constraints between them to ensure problem validity (\cref{fig:teaser}). 
We formalize mathematical properties a well-formed \efa{} must possess as unit tests that can automatically detect violations of these properties. We can then adopt an overgenerate-then-filter approach~\citep{alphacode}, first generating a large number of candidate programs implementing \efa{}s for a seed problem, and then rejecting \efa{}s that fail our tests. 
Finally, we conduct a series of experiments probing properties of the \efa{}s constructed by \method, demonstrating the utility of model-written \efa{}s and testing whether LLMs can be trained to be more successful writers of \efa{}s.

We first show that \efa{}s have properties signaling their coherence. 
\efa{}s are faithful to the seed problem they were derived from: the verifiable problems sampled from an \efa~help a model solve the seed problem the \efa{} was constructed from.
Similarly, the verifiable problems produced by an \efa~are learnable:
when sampling a train and test set from the same \efa{}, a model is able to improve on the test set when given step-by-step solutions of the training problems.

Because \efa{}s allow us to sample a large number of verified problems, we can also use them to 
create more instances of a problem that a model struggles with, or to refresh a static dataset by first constructing an \efa~from a problem that the model already can solve, and then sampling fresh variants using the \efa~that the model struggles with, thereby stress-testing models on similar data.
We show that \method~can be applied to multiple sources of competition-level mathematics problems to automatically construct \efa{}s.
This applicability to multiple kinds of problems allows us to use \efa{}s as a data augmentation for mathematical problem solving on MATH-500~\citep{hendrycksmath2021} and FnEval~\citep{srivastava2024functional}, where we show \efa-based augmentation yields consistent improvements.
Finally, we show that models can improve at inducing \efa{}s from math problems by using the execution feedback from automatic tests in \method~as rewards in a simple reinforced self-training scheme~\citep{zelikman2022star,singh2023beyond,dong2023raft}.

\begin{figure}[t]
    \centering
    \includegraphics[width=1.0\linewidth]{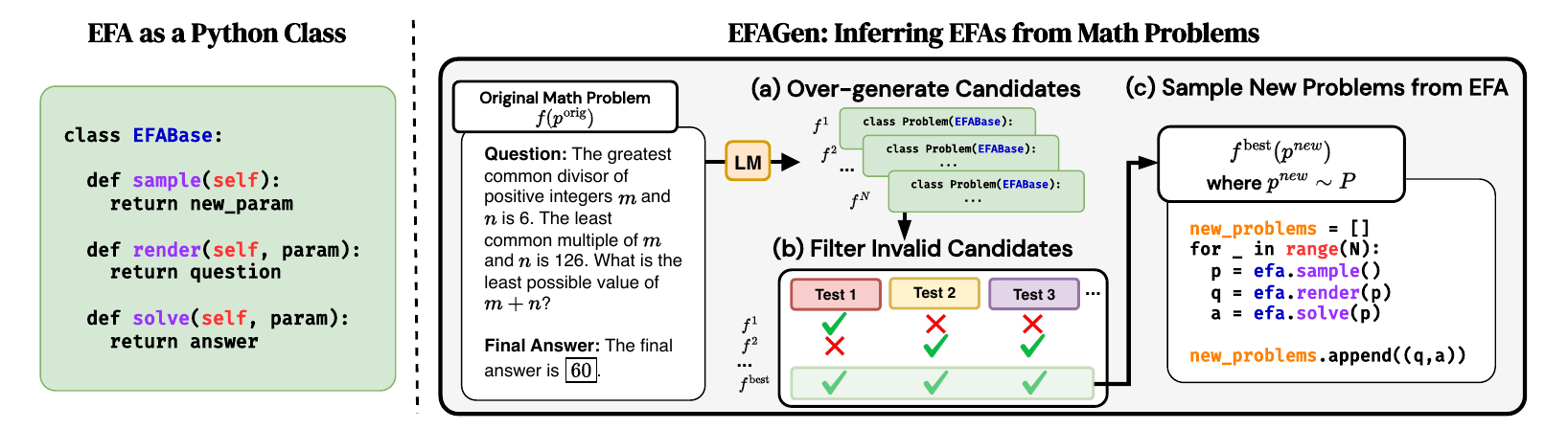}
    \vspace{-1.5em}
    \caption{
    \textbf{Left: Representation of an executable functional abstraction (\efa{})} as a Python class.
    \textbf{Right: Overview of \method{}, a method for automatically inferring \efa{}s} from a math problem.
    In \method{}, we (a) over-generate multiple \efa{} candidates with an LLM and (b) filter out invalid candidates that fail automated tests. The \efa{} can generate new problem variants by sampling parameters and executing the solver. Full code is in \cref{sec:qual-examples}.
    }
    \label{fig:method}
    \vspace{-1em}
\end{figure}

Our contributions in this paper are as follows:
\begin{itemize}[topsep=0pt, itemsep=0.5em, leftmargin=*] 
\item We formalize the notion of Executable Functional Abstractions (\efa{}s) in \cref{sec:efa_structure},  and develop \method{} (\cref{sec:efa_generation}, \cref{fig:method}), an approach that automatically infers \efa{}s from advanced math problems, providing a scalable approach to generate verifiable problem variants with automatic tests for validity and correctness. 
\item We show that these tests can be used as a reward signal for training LLMs to improve at the task of inferring \efa{}s from static problems (\cref{sec:self-improvement}).
\item We show that \method{}~generates faithful (\cref{sec:faithfulness}) and learnable (\cref{sec:learnability}) \efa{}s and can automatically infer \efa{}s from diverse sources of math data (\cref{sec:generalization-numinamath}), and that \efa{}s can be used as a data augmentation (\cref{sec:training-data-augmentation}).

\end{itemize}

\vspace{-0.5em}
\section{Executable Functional Abstractions (\efas)}
\label{sec:method}
\vspace{-0.5em}
Our goal is to automatically convert math problems with static numerical values into \textbf{parameterized abstractions}
that can generate variants of the original problems.
We refer to these parameterized abstractions as \textbf{Executable Functional Abstractions (\efas)}.
\efas{} enable the systematic generation of new problem instances by varying numerical parameters while preserving the underlying problem-solving logic. 
We operationalize the task of inferring an \efa{}~for a static math problem as a program synthesis task where the goal is to write a class implementing the \efa{}.
We use LLMs to generate many candidate \efa{} implementations for a static problem and use a suite of automatic unit tests to filter the candidates by rejecting mathematically unsound ones.
Below,
we describe the desired properties of \efa{}s (\cref{sec:efa_properties}),
how an \efa{} is represented as a Python class (\cref{sec:efa_structure}),
and how we infer \efa{}s from static math problems using LLMs (\cref{sec:efa_generation}).

\vspace{-0.25em}
\subsection{Desired Properties of Abstractions}
\label{sec:efa_properties}
\vspace{-0.25em}

An effective abstraction of a math problem must support variation, preserve validity, and enable automated problem-solving. We identify three core properties of an \efa{}:

\begin{itemize}[nosep,leftmargin=*]
\item \textbf{Structured parameter space}: The abstraction should define a set of parameters that characterize the problem and specify valid relationships among them. This includes identifying which parameters are independent, how dependent parameters are derived, and what constraints must be satisfied to ensure valid problem instances. Such structure enables systematic variation, ensuring that changes to parameters yield meaningful variants with potentially different solutions.
\item \textbf{Procedural generation of instances}: The abstraction should support random sampling of a set of valid parameters (e.g., \texttt{EFA.sample()} in \cref{sec:efa_structure})
and converting the abstract problem into natural language form (e.g., \texttt{EFA.render()} in \cref{sec:efa_structure}),
to help users generate valid problem instances by sampling parameter values within defined constraints.
These constraints are problem-specific and crucial for generating diverse but coherent examples.
\item \textbf{Executable solution logic}: The abstraction should include a method (e.g., \texttt{EFA.solve()} in \cref{sec:efa_structure}) that computes the correct answer for any valid parameter configuration. This solution logic is typically derived from the chain-of-thought~\citep{wei2022chain} used for the static version of the problem and can be implemented as an executable program.
\end{itemize}

\vspace{-0.25em}
\subsection{\efa{} as a Python Class}
\label{sec:efa_structure}
\vspace{-0.25em}
As illustrated in \cref{fig:method} (a), each \efa{} is implemented as a Python class that encapsulates the logic of a math problem in a parameterized form. The class defines a list of parameters along with three key methods:

\begin{itemize}[nosep,leftmargin=*]
    \item \textbf{\texttt{EFA.sample() $\rightarrow$ parameters}}: Samples a valid set of parameters representing problem variants, respecting all constraints specified in the abstraction.
    \item \textbf{\texttt{EFA.render(parameters) $\rightarrow$ question}}: Provides a natural language problem statement, given a specific (sampled) parameter set. This ensures that each generated instance is presented in a format suitable for human or model consumption. In most cases, this involves reusing the problem statement of the seed instruction and mutating the numerical values to be consistent with the given parameters.
    \item \textbf{\texttt{EFA.solve(parameters) $\rightarrow$ answer}}: Computes the correct answer for a given parameter configuration. The solution is expressed as a numerical expression derived through deterministic computations over the parameters. The solver does not need to access the natural language problem statement, as the solution is only dependent on the parameterization of the problem, which is a structured object.
\end{itemize}
These methods operationalize the abstraction and enable automated generation, rendering, and evaluation of math problems.

\vspace{-0.25em}
\subsection{\method{}: Inferring \efa{}s from Math Problems}
\label{sec:efa_generation}
\vspace{-0.25em}
We introduce \method{}, a framework for automatically constructing \efa{}s from static math problems. Given a problem statement and its solution procedure (typically expressed as chain-of-thought reasoning), \method{} uses a large language model (LLM) to generate a candidate \efa{} implementation that captures the logic and structure of the original problem.
This process relies on supervision that is readily available in many math datasets.

Since generating correct and robust code is challenging for LLMs, \method{} adopts an overgenerate-and-filter approach inspired by AlphaCode \citep{alphacode}.
As described in \cref{fig:method}  (a), for each problem, we sample $N$ (e.g., 50) \efa{} candidates and apply a suite of automated tests to discard invalid abstractions. \method{} uses the following tests to validate candidate \efa{}s, as illustrated in \cref{fig:method} (b):

\begin{itemize}[nosep,leftmargin=*]
    \item \texttt{\textbf{is\_extractable(response)}}:
    Verifies that the class contains all required methods.
    
    \item \texttt{\textbf{is\_executable(EFA)}}:  Confirms that the class can be instantiated and executed without errors, and methods like \texttt{EFA.sample()} and \texttt{EFA.solve()} can be called without errors.
    
    \item \texttt{\textbf{has\_dof(EFA)}}: Ensures that sampled parameters differ, rejecting \efas with zero degrees of freedom that cannot produce new problems.
    
    \item \texttt{\textbf{is\_single\_valued(EFA)}}: Confirms that identical parameters yield equivalent solutions, rejecting impermissible implementations including multivalued functions or logically incoherent abstractions.
    
    \item \texttt{\textbf{matches\_original(EFA, orig\_params, orig\_sol)}}:  Validates that the abstraction, when instantiated with the original parameters, produces the original problem and solution. This serves as a cycle-consistency or soundness check.
\end{itemize}

Any program that fails these tests cannot logically be a valid implementation of an \efa{}.
\method{} enables generation of \efa{}s at scale, as shown in \cref{fig:method} (c), as large numbers of candidate \efa{}s can be generated and filtered automatically.
Over time, these tests can also be used to fine-tune LLMs toward better abstraction generation, such as with reinforced self-training \citep{singh2023beyond,dong2023raft} or reinforcement learning with verifiable rewards \citep{lambert_tulu_2024}.

\vspace{-0.25em}
\section{Experiments \& Results}
\label{sec:experiments}
\vspace{-0.25em}

\paragraph{Datasets.} Throughout this section, we use the following datasets in our experiments:
\begin{itemize}[topsep=0pt,noitemsep,leftmargin=*]
    \item \textbf{MATH}~\citep{hendrycksmath2021}. Competition math dataset with a test set that consists of 5000 math problems described in text comprising different categories and five levels of difficulty. As we will show in \cref{sec:self-improvement}, LLMs struggle with task of \efa{} generation and therefore, we improve their performance by training on the \efa{} generation task using the MATH train set consisting of 7500 problems (similar distribution as the test set). 
    \item \textbf{MATH-Hard.} With our focus on challenging math problems, this is a subset of MATH test problems of the highest difficulty (level 5) across all categories (1387 problems). 
\end{itemize}

\vspace{-0.25em}
\paragraph{Metrics.} To evaluate the performance of LLMs we use the following metrics:
\begin{itemize}[topsep=0pt,noitemsep,leftmargin=*]
\item \textbf{EFA Success Rate.} We measure the ability of LLMs to generate valid, high-quality EFAs (defined in \cref{sec:efa_properties}) as the frequency (\%) of EFAs generated that past all the diagnostic tests outlined in \cref{sec:efa_generation}.
\item \textbf{Pass$@k$ Rate (\%).} Following \citet{chen2021evaluating}, we measure the ability of LLMs to solve math problems by sampling 25 generations with temperature sampling and estimating the unbiased pass$@k$ rate, i.e., the likelihood that out of $k$ generated solutions any one corresponds to the correct answer.
\end{itemize}

\vspace{-0.25em}
\subsection{Self-Improvement: LMs Can Improve at Inferring \efa{}s With Execution Feedback}
\label{sec:self-improvement}
\vspace{-0.25em}

\begin{figure}
    \centering
    \includegraphics[width=\linewidth]{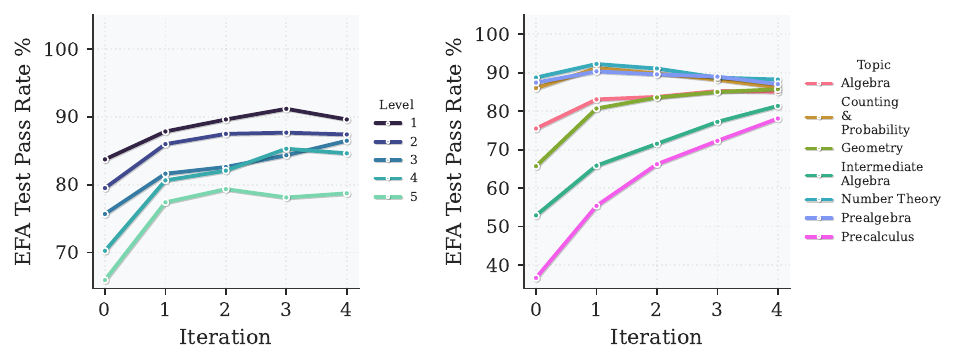}
    \vspace{-1.5em}
    \caption{\textbf{LLMs can use our tests to self-improve at inferring \efa{}s.} We plot the percentage of constructed \efa{}s passing all tests across iterations of self-training, grouped by MATH problem difficulty (left) and by problem category (right). 
    Harder difficulty levels and problem categories are harder to infer \efa{}s for and improve more during training.
    }
    \label{fig:iterations}
    \vspace{-0.5em}
\end{figure}

Inferring valid \efa{}s across diverse math problems is challenging, especially as the difficulty and complexity of topics increases. For instance, as shown in \cref{fig:iterations}, \llama{}~\citep{Team2024TheModels} struggles to generate valid \efa{}s for Level 5 problems and for topics such as Precalculus in the MATH dataset, where it is only able to infer valid \efa{}s for $\approx\!35\%$ of Precalculus questions. 
In \cref{sec:method}, we introduce a number of unit tests (i.e., verifiable rewards) that indicate whether a generated \efa{} is valid.
Here, we show that we can train models to improve on inferring valid \efa{}s by self-training according to these tests. 
Specifically, we use a rejection-finetuning approach~\citep{zelikman2022star, singh2023beyond, dong2023raft}, in which we sample \efa{} candidates from a model and filter according to our rewards to construct a training dataset of correct examples. 
We begin with the MATH training set (7,500 problems) and sample 10 candidate \efa{}s per problem. Candidates failing any of the reward checks are discarded. The remaining valid examples form a dataset for supervised fine-tuning. This process -- sampling, filtering, and retraining -- is repeated over 5 iterations (see \cref{sec:training-details} for details).

We report the EFA success rates across iterations in \cref{fig:iterations}, where we group by difficulty levels (left) and by annotated problem category (right). 
Success rates steadily improve over training iterations, especially for harder problems. 
At iteration 0 (before training), we observe that harder problems (e.g., Level 5) are also harder to infer \efa{}s for, with EFA success rates $\approx 17\%$ lower for Level 5 than Level 1 problems.
Similarly, certain categories like `\textit{Intermediate Algebra}', `\textit{Counting}' and `\textit{Probability}' are harder to infer \efa{}s for. 
These domains generally see the most significant increases from training. 
Between iteration 1 and iteration 5, the Intermediate Algebra's EFA success rate showed the most significant increase, rising from 52.93\% to 81.38\%, and Geometry improved from 65.71\% to 85.71\%. Additionally, the pass rate for Level 5 problems increased from 65.95\% to 78.73\%. These changes indicate substantial improvements in the model's ability to infer \efa{} across these dataset slices.
The final model trained for 5 iterations becomes the basis for our \method{} method. 

\subsection{Faithfulness: \efa{}s Capture the Reasoning Required to Solve the Seed Problem}
\label{sec:faithfulness}

To evaluate the faithfulness of \efa{}s, we ask: can the generated variant problems improve a model's solve rate on the original seed problem?
We select all of problems from MATH-Hard for which \llama{}'s pass@5 rate $< 50\%$ and for which \method{} can successfully infer an \efa{} using the gold solution.\footnote{Based on the intuition that testing for faithfulness requires an \efa{} (i.e., requires a problem that can be solved in principle) but improving requires a problem that is not solved 100\% of the time.}
For each problem, we sample additional problem variants (we ensure their parameters differ from the seed problem) until \llama{}~solves one correctly. 
We then check if \llama{} can solve the original problem, given the variant and its solution as an in-context example.
Results in \cref{tab:faithfulness-learnability} (left) show a 23.07\% absolute improvement in pass@1 rate, indicating that \efa{}-generated variants
can teach model the problem-solving reasoning needed for the seed problem.

\vspace{-0.2em}
\subsection{Learnability: Performance on Generated Problems Should Increase with Experience}
\label{sec:learnability}
\vspace{-0.2em}

An effective problem abstraction should enable a model to solve both the original seed problem and its variants. To evaluate this, we test whether training on \efa{}-generated problem variants helps a model solve additional variants that are drawn from the same \efa{} but are different from the seed problem.

We sample 1,000 \efa{}s inferred from the MATH-Hard test set and generate one new variant per \efa, forming a held-out test set. 
For each \efa, we also identify one variant that \llama{} solves correctly.
We then test \llama{}’s performance on the held-out test set, with and without access to that solved variant as an in-context example.
As shown in \cref{tab:faithfulness-learnability}, access to one correctly-solved variant improves the model’s pass rate on other variants by 16.65\% on average. This demonstrates that reasoning learned from one variant reliably transfers to others within the same abstraction.
\begin{table}[t]
\centering
\resizebox{\linewidth}{!}{
\begin{tabular}{ccccccc}
\toprule
                     \multicolumn{3}{c}{\textbf{Faithfulness (\cref{sec:faithfulness}): \efa helps on the original problem}} & \multicolumn{3}{c}{\textbf{Learnability (\cref{sec:learnability}): \efa~helps on its variants}} \\ 
\cmidrule(l{0.5em}r{0.5em}){1-3} \cmidrule(l{0.5em}r{0.5em}){4-6}
                     Initial Pass@1        & +Data from \efa        &  Sample Size        & Initial Pass@1        & +Data from \efa        & Sample Size \\
                     \midrule
                     15.66          & 38.73 \plus{23.07}         & 307          & 14.58          & 31.23 \plus{16.65}         & 1,000          \\
                      \bottomrule
\end{tabular}
}
\vspace{-0.25em}
\caption{\textbf{\efa{}s faithfully capture the solutions of the problems they were derived from (left), and problem variants constructed by \efa{}s share learnable structure (right).} 
Left: Giving solutions to problems variants from an \efa~ as in-context examples nearly doubles the solve rate of an LLM on the seed problem the \efa~was derived from. 
Right: Giving solutions to problem variants from an \efa{} as in-context examples helps an LLM solve a holdout set of variants from the same \efa.
See \cref{sec:faithfulness} and \cref{sec:learnability} for details.
}
\label{tab:faithfulness-learnability}
\end{table}

\vspace{-0.2em}
\subsection{Augmentation: \efa{}s Are Effective at Expanding Static Math Datasets}
\label{sec:training-data-augmentation}
\vspace{-0.2em}

While high-quality math datasets exist, these are often expensive to construct.
\method{} offers a scalable solution by generating diverse, faithful problem variants through \efa{}s, thereby augmenting existing datasets.
To demonstrate this, we fine-tune \llamabase{} using \efa{}-generated data derived from the MATH training set.
Concretely, we annotate 7,500 training problems with step-by-step reasoning from a teacher model (\llama{}). We ensure that the reasoning is correct by filtering out the reasoning that yields incorrect answers.
Then, for each of the 7,500 problems, we construct an \efa{} and sample one problem variant.
We compare two training settings.
In the first setting, we use only the teacher-labeled seed data.
In the second, we augment the seed data by adding \efa{}-generated examples in a 1:1 ratio.
We perform experiments with both 33\% (2,500) and 100\% (7,500) of the seed data and evaluate performance on three benchmarks: MATH-500 split~\citep{lightman2023let} and the November and December splits of FnEval, each containing perturbed versions of MATH problems.
See \cref{sec:math-training} for hyperameter details.

As shown in \cref{tab:augmentation_improve_performance}, \efa{}-based augmentation leads to consistent improvements across all evaluation metrics: Pass@1, Pass@10 rate, and Majority@25~\citep{Wang2022Self-ConsistencyModels}. For example, in the 33\% seed setting, Pass@1 improves by +1.9 on MATH-500 and by +2.2 on both FnEval splits.
In the full 100\% seed setting, the gain still holds, underscoring the value of \efa{}s in enhancing data quality and model performance.

\begin{table}[t]
\resizebox{\linewidth}{!}{
\begin{tabular}{@{}l ccc ccc ccc@{}}
\toprule
& \multicolumn{3}{c}{MATH-500} & \multicolumn{3}{c}{FnEval (November Split)} & \multicolumn{3}{c}{FnEval (December Split)} \\
\cmidrule(l{0.5em}r{0.5em}){2-4} \cmidrule(l{0.5em}r{0.5em}){5-7} \cmidrule(l{0.5em}r{0.5em}){8-10}
Training Data & Pass @ 1 & Pass @ 10 & Maj @ 25 & Pass @ 1 & Pass @ 10 & Maj @ 25 & Pass @ 1 & Pass @ 10 & Maj @ 25 \\
\midrule
MATH (33\%) & 22.4 & 56.4 & 36.8 & 24.5 & 55.3 & 39.6 & 24.4 & 55.4 & 39.3 \\
+\efa (1:1) & 24.3 & 58.3 & 38.8 & 26.7 & 59.2 & 41.8 & 26.6 & 57.3 & 41.2 \\
& \plus{1.9} & \plus{1.9} & \plus{2.0} & \plus{2.2} & \plus{3.9} & \plus{2.2} & \plus{2.2} & \plus{1.9} & \plus{1.9} \\
\midrule
MATH (100\%) & 24.3 & 57.8 & 37.0 & 26.8 & 58.6 & 43.1 & 26.5 & 57.6 & 41.5 \\
+\efa (1:1) & 26.1 & 60.6 & 40.4 & 29.3 & 60.1 & 44.3 & 28.8 & 59.6 & 43.7 \\
& \plus{1.8} & \plus{2.8} & \plus{3.4} & \plus{2.5} & \plus{1.5} & \plus{1.2} & \plus{2.3} & \plus{2.0} & \plus{2.2} \\
\bottomrule
\end{tabular}
}
\vspace{-0.25em}
\caption{\textbf{\efa{}-based data augmentation is consistently effective.} Comparison with and without synthetic data augmentation using problems drawn from generated \efa{}s. 
The table shows performance across MATH-500 and FnEval benchmarks (November and December snapshots). When augmenting, we use a 1:1 ratio of examples drawn from training data vs. from an \efa{}, and report results using 33\% of the MATH train set and 100\% of the train set. 
}
\label{tab:augmentation_improve_performance}
\vspace{-1em}
\end{table}

\vspace{-0.25em}
\subsection{\method{} Complements Existing Synthetic Data Generation Approaches}
\label{sec:synthetic-data-complement}
\vspace{-0.25em}

\efa{}s are designed to complement, not replace, existing synthetic data generation approaches. 
To demonstrate this complementary relationship, we conduct experiments with high-quality synthetic data from NuminaMath~\citep{li2024numinamath}, which aggregates synthetic data from various sources, showing that \method{} can infer \efa{}s for synthetic data and use these \efa{}s to augment synthetic datasets at different scales.

We sample 1k, 2.5k, and 5k problems with step-by-step solutions from the \texttt{synthetic\_math} and \texttt{synthetic\_amc} sources in NuminaMath. 
For each sample, we apply \method{} to infer \efa{}s, generate one problem variant from each \efa{}, and use rejection sampling to create training data from the \efa{}s. 
We train three models at each scale: one trained only on the NuminaMath synthetic data (\textit{NuminaMath Synthetic}), one trained only on data derived from \efa{}s (\textit{EFA Generated}), and one trained on the NuminaMath synthetic data augmented with our \efa{}-derived data (\textit{NuminaMath Synthetic + EFA Generated}).

Results on MATH-500 are shown in \cref{tab:synthetic-data-complement}.
At each scale, the model trained on synthetic data augmented with \efa{}-generated data performs best across most metrics. 
Notably, the \efa{}-generated data typically outperforms the original synthetic NuminaMath data, suggesting that the \efa{} inference process produces high-quality problem variants that enhance model learning.
These results demonstrate that \method{} provides a scalable approach for augmenting existing synthetic datasets, effectively complementing current synthetic data generation methods.

\begin{table}[t]
\centering
\resizebox{\linewidth}{!}{
\begin{tabular}{@{}l l cccc@{}}
\toprule
& & \multicolumn{4}{c}{\textbf{MATH-500 Performance}} \\
\cmidrule(l{0.5em}r{0.5em}){3-6}
\textbf{Scale} & \textbf{Data Mix} & \textbf{Pass@1} & \textbf{Pass@5} & \textbf{Pass@10} & \textbf{MV Acc} \\
\midrule
\multirow{3}{*}{1k} & NuminaMath Synthetic & 20.8 & 45.6 & 56.4 & \textit{38.6} \\
& \efa{} Generated & \textit{24.0} & \textit{48.5} & \textbf{58.7} & \textit{38.6} \\
& NuminaMath Synthetic + \efa{} Generated & \textbf{24.4} & \textbf{48.5} & \textit{58.2} & \textbf{40.6} \\
& & \plus{3.7} & \plus{2.9} & \plus{1.8} & \plus{2.0} \\
\midrule
\multirow{3}{*}{2.5k} & NuminaMath Synthetic & 23.0 & \textit{47.6} & \textit{58.5} & \textit{38.8} \\
& \efa{} Generated & \textit{23.1} & 47.0 & 57.2 & 35.8 \\
& NuminaMath Synthetic + \efa{} Generated & \textbf{24.9} & \textbf{50.5} & \textbf{61.1} & \textbf{41.6} \\
& & \plus{1.9} & \plus{2.9} & \plus{2.6} & \plus{2.8} \\
\midrule
\multirow{3}{*}{5k} & NuminaMath Synthetic & 20.9 & 46.3 & 57.0 & \textit{39.8} \\
& \efa{} Generated & \textit{23.6} & \textit{48.6} & \textit{59.2} & \textit{39.8} \\
& NuminaMath Synthetic + \efa{} Generated & \textbf{26.7} & \textbf{51.9} & \textbf{62.1} & \textbf{44.0} \\
& & \plus{5.8} & \plus{5.6} & \plus{5.0} & \plus{4.2} \\
\bottomrule
\end{tabular}
}
\vspace{-0.25em}
\caption{\textcolor{black}{\textbf{\method{} complements existing synthetic data generation approaches.} 
Performance comparison across different data scales (1k, 2.5k, 5k) when training models on: NuminaMath synthetic data alone, \efa{}-generated data alone, and both combined. 
The combined approach typically performs best, with \efa{}-generated data generally outperforming the original synthetic data. 
The \plus{} values show absolute improvements over the NuminaMath Synthetic baseline within each scale.
1st-place is \textbf{bold}, 2nd is \textit{italicized}.}
}
\label{tab:synthetic-data-complement}
\vspace{-1em}
\end{table} 

\vspace{-0.25em}
\subsection{Quality Analysis: Low-Quality \efa{}s Are Naturally Filtered Out}
\label{sec:efa-quality-analysis}
\vspace{-0.25em}

A potential concern with \method{} is that the automated \efa{} generation process may produce low-quality abstractions that could negatively impact training. To address this, we analyze how rejection sampling naturally filters out problematic \efa{}s during the training data generation process.

We identify ``bad'' \efa{}s using an LLM with heuristics that flag abstractions exhibiting common failure modes: trivial problems, extraneous variables, or hard-coded values. We then compare the training data yield rates (the percentage of responses that receive non-zero rewards during rejection sampling) between good and bad \efa{}s.

As shown in \cref{tab:efa_quality_yield}, low-quality \efa{}s have significantly lower yield rates compared to good \efa{}s. With a single answer attempt, bad \efa{}s contribute training data only 5.04\% of the time, compared to 27.0\% for good \efa{}s -- a ratio of over 5 to 1 in favor of good data. Even when allowing up to 5 answer attempts, the ratio remains favorable at 4.51 to 1. This demonstrates that as long as rejection sampling or reinforcement learning is used, noisy \efa{}s naturally filter themselves out, ensuring that good data significantly outnumbers bad data in the final training set.

To further validate the quality of \efa{}-generated data, we conduct a direct comparison between training exclusively on problem variants generated by \efa{}s versus training exclusively on real problems from the MATH training set. As shown in \cref{tab:real_vs_synthetic_comparison}, despite potential noise in rejection-sampled \efa{} data, models trained on synthetic data achieve nearly identical performance to those trained on real data (22.6\% vs 22.4\% Pass@1 on MATH-500). This shows that \efa{}-generated data is as effective as existing math data for model training.

\begin{table}[t]
\centering
\resizebox{\linewidth}{!}{
\begin{tabular}{@{}l cc c@{}}
\toprule
& Good \efa{}s & Bad \efa{}s & Good to Bad Data Ratio \\
\midrule
Training Data Yield Rate (1 Answer Attempt) & 27.0\% & 5.04\% & 5.36 to 1 \\
Training Data Yield Rate (5 Answer Attempts) & 39.9\% & 8.85\% & 4.51 to 1 \\
\bottomrule
\end{tabular}
}
\vspace{-0.25em}
\caption{\textcolor{black}{\textbf{Low-quality \efa{}s are naturally filtered out during rejection sampling.} We compare the training data yield rates (percentage of responses that receive non-zero rewards) between good and bad \efa{}s. Bad \efa{}s are identified using LLM-based heuristics that flag trivial problems, extraneous variables, or hard-coded values. The low yield rates of bad \efa{}s mean they contribute minimally to training data.}}
\label{tab:efa_quality_yield}
\vspace{-1em}
\end{table}

\begin{table}[t]
\centering
\resizebox{\linewidth}{!}{
\begin{tabular}{@{}l ccc ccc ccc@{}}
\toprule
& \multicolumn{3}{c}{MATH-500} & \multicolumn{3}{c}{FnEval (November)} & \multicolumn{3}{c}{FnEval (December)} \\
\cmidrule(l{0.5em}r{0.5em}){2-4} \cmidrule(l{0.5em}r{0.5em}){5-7} \cmidrule(l{0.5em}r{0.5em}){8-10}
Training Data & Pass @ 1 & Pass @ 10 & Maj @ 25 & Pass @ 1 & Pass @ 10 & Maj @ 25 & Pass @ 1 & Pass @ 10 & Maj @ 25 \\
\midrule
Real Data Only & 22.4 & 56.4 & 36.8 & 24.4 & 55.4 & 39.3 & 24.5 & 55.3 & 39.6 \\
Synthetic Data Only & 22.6 & 58.0 & 37.8 & 24.9 & 56.6 & 38.3 & 25.5 & 57.2 & 40.0 \\
\bottomrule
\end{tabular}
}
\vspace{-0.25em}
\caption{\textcolor{black}{\textbf{\efa{}-generated data performs comparably to real data.} Direct comparison of training exclusively on problem variants generated by \efa{}s versus training exclusively on real problems from the MATH training set. Despite potential noise in rejection-sampled \efa{} data, models trained on synthetic data achieve nearly identical performance to those trained on real data.}}
\label{tab:real_vs_synthetic_comparison}
\vspace{-1em}
\end{table}

\vspace{-0.25em}
\subsection{Generality: \method{} Can Work Across Diverse Math Domains}
\label{sec:generalization-numinamath}
\vspace{-0.25em}

\begin{figure}
    \centering
    \includegraphics[width=0.9\linewidth]{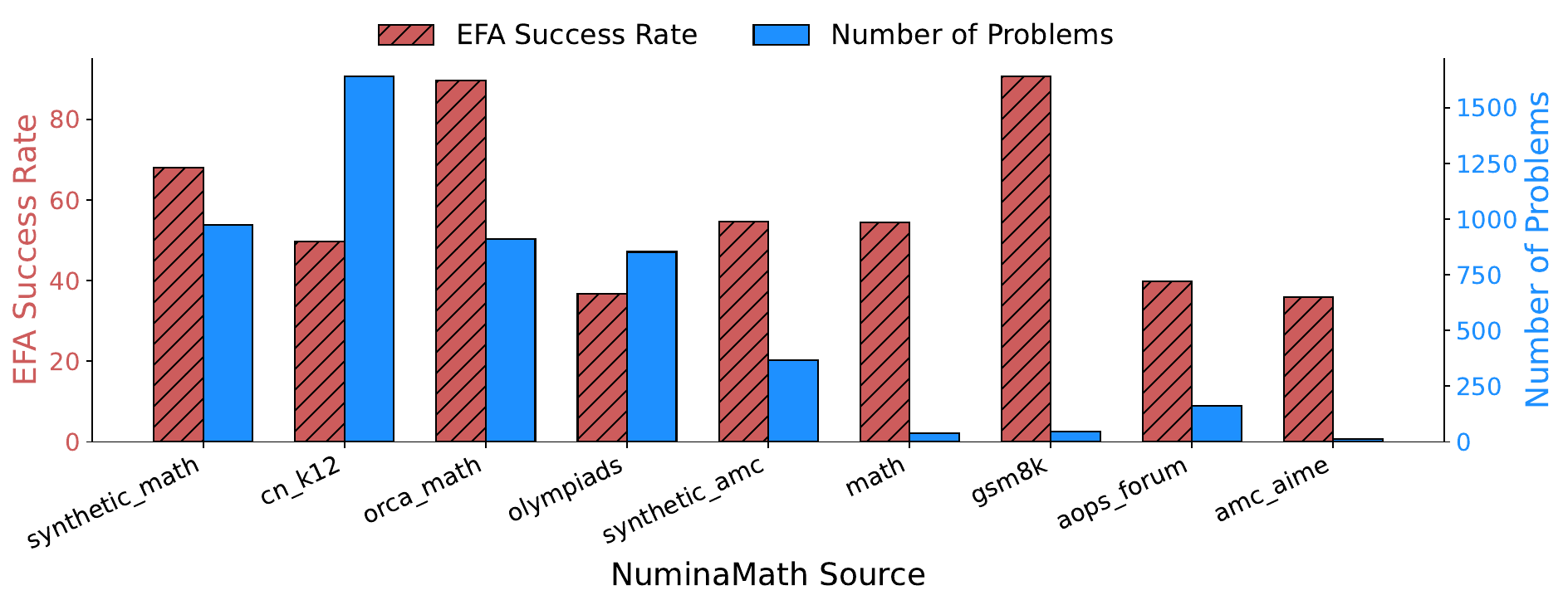}
\vspace{-0.5em}
\caption{\textbf{\method{} can infer \efa{}s for diverse sources of math problems.} Here, we show the results of applying \method{} to infer \efa{}s for the NuminaMath~\citep{li2024numinamath} dataset, which contains a mix of math problems from a diversity of sources ranging from grade school mathematics (GSM8K) to
national/international olympiads (olympiads). \method{} achieves a nonzero success rate across all sources of problems.
}
\label{tab:functionalization-success-rates-numinamath}
\end{figure}

Importantly, \method{} generalizes beyond the distribution of questions in the MATH dataset. As detailed in \cref{tab:functionalization-success-rates-numinamath}, our approach successfully infers \efa{}s across various math sources from the NuminaMath dataset \citep{li2024numinamath} -- ranging from grade-school problems (GSM8K) to national/international competitions (e.g., AMC, AIME, IMO). This demonstrates the broad applicability of \efa{}s for structuring and scaling math data across diverse domains.
We generally see that easier math domains like GSM8K are easier to infer \efa{}s for than harder domains like AIME or Olympiad problems; nevertheless, \method{} can infer some successful \efa{}s even on the hardest domain.

\textcolor{black}{To further demonstrate the scalability of \method{}, we evaluate its performance on a larger set of 10,000 competition-level math problems from NuminaMath. As shown in \cref{tab:numinamath-competition}, we are able to successfully infer \efa{}s at rates of 38.4\%, 50.9\%, and 40.6\% for the Olympiads, Synthetic AMC, and AMC-AIME sources in NuminaMath, respectively. The 95\% confidence intervals for each source are significantly above 0\% (the lowest is 33.7\%), demonstrating that \method{} can reliably infer \efa{}s for the hardest problems in large math training datasets.}

\textcolor{blue}{\begin{table}[t]
\centering
\resizebox{0.7\linewidth}{!}{
\begin{tabular}{@{}lccc@{}}
\toprule
\textbf{Source} & \textbf{Functionalization Rate (\%)} & \textbf{95\% CI (\%)} & \textbf{Num Problems} \\
\midrule
Olympiads & 38.4 & [37.2\%, 39.5\%] & 6,950 \\
Synthetic AMC & 50.9 & [49.0\%, 52.7\%] & 2,881 \\
AMC-AIME & 40.6 & [33.7\%, 48.4\%] & 169 \\
\bottomrule
\end{tabular}
}
\vspace{-0.25em}
\textcolor{black}{\caption{\textbf{\method{} can infer \efa{}s for large-scale competition-level mathematics.} 
Across 10,000 competition-level problems in NuminaMath, we successfully infer \efa{}s at substantial rates across different sources. The 95\% confidence intervals are significantly above 0\% (lowest is 33.7\%), demonstrating that \method{} can reliably infer \efa{}s for the hardest problems available in large math training datasets.}
\label{tab:numinamath-competition}}

\end{table} }

\section{Related Work}
\label{sec:related_work}

\paragraph{Symbolic Approaches to Math Reasoning.} 
A distinct line of prior work has focused on assessing the true mathematical reasoning capabilities of LLMs, specifically by measuring the ``reasoning gap'' or the drop in math reasoning performance after perturbing questions in existing datasets~\citep{shi2023large, zhou2025gsm, huang2025math, ye2025physics}. One prominent approach is to generate different or difficult math questions conditioned on an existing question but test skills by employing frontier models~\citep{zhang2024darg, patel2025get} or human annotators~\citep{srivastava2024functional, shah2024aiassisted, huang2025math}. For instance, \citet{srivastava2024functional} propose FnEval dataset by manually functionalizing select problems from the MATH dataset~\citep{hendrycksmath2021} that can be subsequently used to sample multiple distinct math problems testing similar skills (albeit with different numerical variables). Similarly, \citet{mirzadeh2025gsmsymbolic} release the \texttt{GSM-Symbolic} dataset that augments the existing GSM8K dataset~\citep{cobbe2021gsm8k} with templates containing placeholders for several numeric and textual variables and can be used to sample distinct math word problems for a robust evaluation of LLM's reasoning abilities. In contrast, to this line of work requiring expensive annotations from humans or frontier models (thereby, hindering scalability) and tailored to specific, predefined math datasets; we propose \method{} that automatically functionalizes \emph{any} math problem using relatively small language models making it \textit{widely-applicable} and \textit{scalable}, i.e., able to sample a potentially infinite number of related math problems from any distribution or dataset. Moreover, the aforementioned prior work only focuses on the robust evaluation of LLMs, whereas we extend the concept of abstraction for downstream applications via training, as shown in~\cref{sec:training-data-augmentation}.

\paragraph{Data and Environment  Generation.} 
Past work has generally approached improving models on reasoning tasks like math by generating large amounts of broad-coverage training data.
This trend builds on work in generating instruction-tuning data \citep{wang2023self}, where model-generated instructions have been used to teach models to follow prompts.
\citet{luo2023wizardmath} introduced generation method based on Evol-Instruct \citep{xu2023wizardlm}, which augmented a seed dataset of math problems by generating easier and harder problems.
Related lines of work have sought to expand datasets by augmenting existing math datasets \citep{yu2024metamath}, adding multiple reasoning strategies \citep{yuemammoth}, covering challenging competition problems \citep{li2024numinamath}, or curating responses \citep{liu2024acemath}. 
The data generated in these settings differs from our data in a number of respects: first, it is generally broad-coverage, focusing on large-scale diverse data, as opposed to targeted, instance-specific data. 
This direction was also explored by \citet{khandataenvgym}, who define data generation agents that can generate specific data based on a particular model's weaknesses, covering math and several other domains. 
Finally, past work that has augmented a seed dataset (e.g., \citet{yu2024metamath, yuemammoth}) has done so by modifying problems in the surface form, whereas our method first infers a latent structure and then creates problems by sampling from the structure. 
In contrast, \method{} focuses on generating similar examples of existing data by inferring an underlying structure from an example; we show that this has applications to data generation for augmentation but also for stress-testing or measuring the performance gap of models on similar problems.

\vspace{-0.5em}
\section{Conclusion}
\label{sec:conclusion}
\vspace{-0.5em}

We introduce Executable Functional Abstraction (\efa{}), a math abstraction that encapsulates the logic of a math problem in a parameterized form, enabling the automated sampling of variant problems.
Building on this definition, we propose \method, a framework that infers \efa{}s via program synthesis using large language models (LLMs) that we train on easy-to-compute \efa{} rewards. 
Concretely, our approach uses an LLM to over-generate \efa{} candidates, which are then filtered using a suite of diagnostic tests that verify their validity. We demonstrate that \method{} can successfully infer \efa{}s from diverse math problems—and that incorporating execution feedback as a reward in a simple self-training scheme further improves its performance. Moreover, models trained on \efa{}-based math problems not only perform better on the generated variants but also improve accuracy on the original seed problems. Finally, we show that \efa{}s provide a scalable solution for augmenting diverse problem variants across various math datasets.

\section*{Acknowledgments}
This work was supported by DARPA ECOLE Program No. HR00112390060,
NSF-CAREER Award 1846185, NSF-AI Engage Institute DRL-2112635, DARPA Machine Commonsense (MCS) Grant N66001-19-2-4031, ARO Award W911NF2110220, ONR Grant N00014-23-1-2356, Microsoft Accelerate Foundation Models Research (AFMR) grant program, and a Bloomberg Data Science PhD Fellowship. The views contained in this article are those of the authors and not of the funding agency.

\bibliography{colm2025_conference}
\bibliographystyle{colm2025_conference}

\appendix
\section*{Appendix}
\textcolor{black}{
The section Adversarial Search (\cref{fig:hard-variant-search}) outlines how \efa{}s can generate challenging problem variants to probe model weaknesses. The Scaling section (\cref{sec:scaling}) investigates the effect of the number of sampled variants per \efa{}, showing how performance trends with increased augmentation. The Ablation section (\cref{sec:unit-tests-ablation}) analyzes the impact of applying unit tests during \efa{} generation on downstream data quality. Qualitative Examples (\cref{sec:qual-examples}) presents representative \efa{}s spanning several MATH domains, including algebra, number theory, and probability, illustrating the range and structure captured by the method. The Experimental Details section describes all data generation, augmentation, and model training settings—\efa{} generation (\cref{efagen-prompt}), rejection finetuning and variant sampling protocols (\cref{sec:training-details}), math inference configuration, and details for math-specific training (\cref{sec:math-training}).}
\begin{figure}
    \centering
    \includegraphics[width=1.0\linewidth]{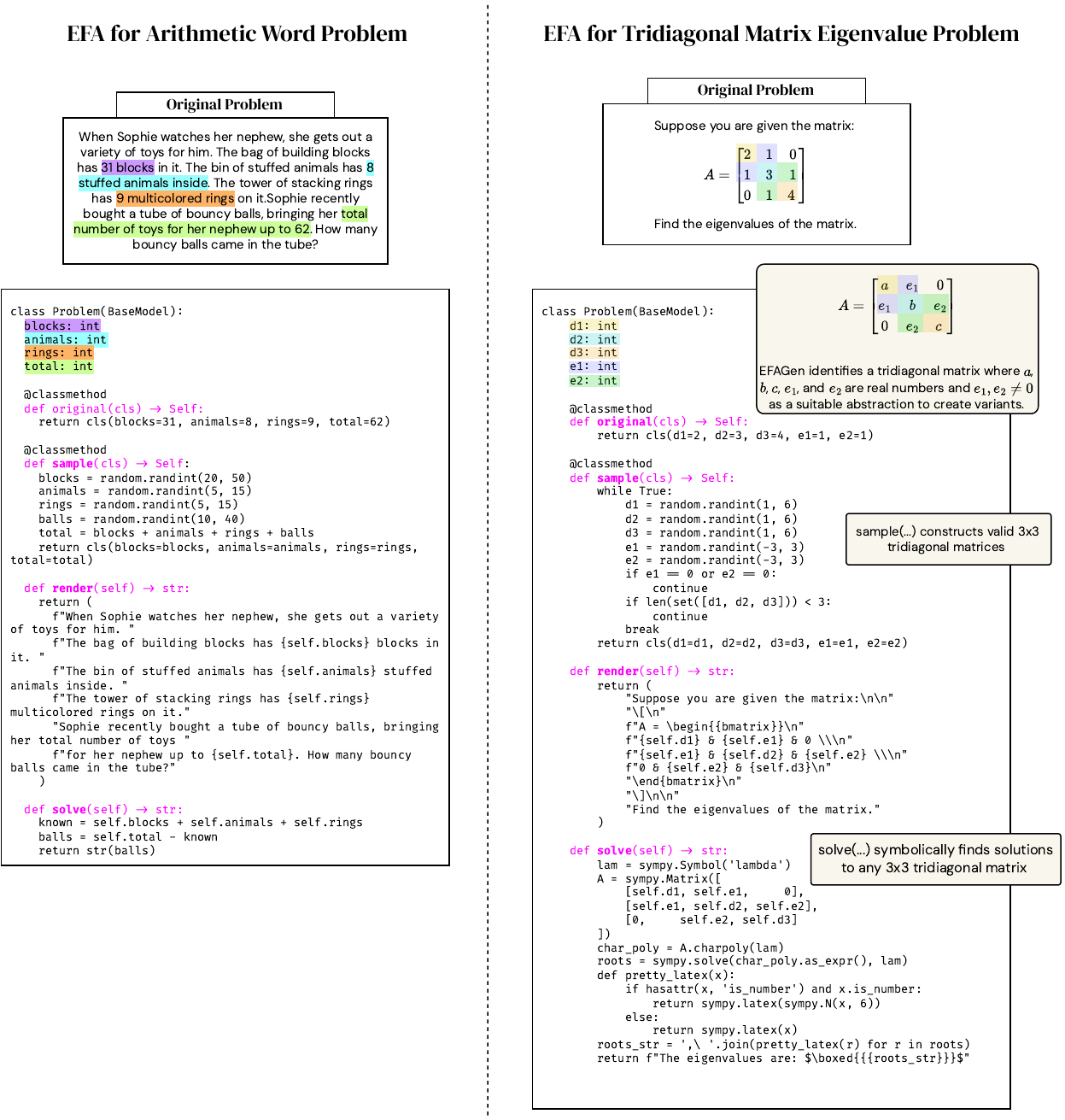}
    \caption{\efa{}s inferred for problems shown in \cref{fig:efa-vs-template-based}. On the left is an \efa{} for a grade-school level math word problem. On the right is an \efa{} for the tridiagonal matrix eigenvalue problem. \efa{}s are able to represent both types of problems, despite the wide gap in problem complexity. The \texttt{sample} method constructs mathematical objects with required properties, while the \texttt{solve} method implements a generalized solution for any object constructible by the \texttt{sample} method. See \cref{sec:efa_structure} for a more detailed explanation.}
    \label{fig:expanded-efas-for-arithmetic-and-tridiagonal}
\end{figure}
\section{Adversarial Search: \method{} Can Find Hard Problem Variants}
\begin{figure}
    \centering
    \includegraphics[width=0.5\textwidth]{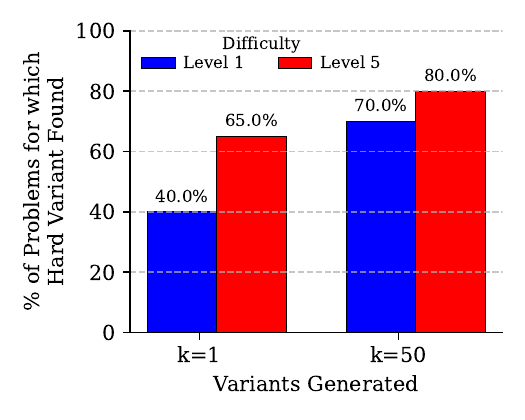}
    \caption{\textbf{\efa{}s can find harder variants of problems.} We infer an \efa{} for a sample of Level 1 (easiest) and Level 5 (hardest) seed problems \gpt~solves correctly, and generate $k$ variants of each problem. We plot the percentage of seed problems for which a variant that \gpt~solved incorrectly was found.
    }
    \label{fig:hard-variant-search}
\end{figure}
\efa{}s can also be used for evaluation or as a source of targeted training data by finding hard instances that models struggle with. 

To demonstrate this, we randomly sample problems from the MATH training that are correctly solved by a strong model (\gpt{}); we sample $N=20$ of both Level 1 (easiest) and Level 5 (hardest) problems.
For each problem, we construct an \efa{} using \method{} and then sample $50$ variants from the \efa{}.
We attempt to solve each variant with \gpt, and measure for what fraction of problems we are able to find variants among the $50$ samples that \gpt~cannot solve.
This is an estimate of the probability that we can use an \efa{} to sample problems that cannot be solved by the model, even when the seed problem is solvable. 
The results are shown in \cref{fig:hard-variant-search} where we see that there is a non-zero probability of finding hard variants to a given problem, even for easy problems (i.e., Level 1 in MATH) and with a strong model like \gpt{}. 

\textcolor{black}{\section{Scaling: \method{} Scales Effectively Up to 16 Examples per \efa{}}}
\label{sec:scaling}

\textcolor{black}{To understand the scaling behavior of \efa{}-based data augmentation, we investigate how performance varies with the number of problem variants generated per \efa{}. 
We sample 100 unique \efa{}s from the MATH training set and vary the number of problem variants generated by each \efa{} from 1 to 64. 
For each scaling setting, we train \llamabase{} on the generated data and evaluate on MATH-500.}

\textcolor{black}{As shown in \cref{tab:scaling_examples}, we observe smooth scaling improvements as we increase the number of variants from 1 to 16 examples per \efa{}, with performance gains plateauing beyond 16 examples. 
Specifically, Pass@1 improves from 14.1\% with 1 example per \efa{} to 23.8\% with 16 examples, while Pass@10 increases from 48.5\% to 57.6\% over the same range. 
However, scaling begins to saturate at 32 and 64 examples per \efa{}, suggesting that sampling too many problem variants from each \efa{} uniformly may hurt diversity and lead to diminishing returns.
The optimal scaling point appears to be around 16 examples per \efa{}, where three of the four metrics achieve their peak performance.}

\begin{table}[t]
\centering
\resizebox{.8\linewidth}{!}{
\begin{tabular}{@{}c cccc@{}}
\toprule
Training Data per \efa & Pass@1 & Pass@5 & Pass@10 & Majority Vote Accuracy \\
\midrule
1 & 14.1 & 37.2 & 48.5 & 29.6 \\
2 & 19.1 & 42.8 & 53.3 & 34.0 \\
4 & 21.9 & 45.1 & 54.7 & 35.4 \\
8 & 22.9 & 46.9 & 57.4 & 35.6 \\
16 & 23.8 & \textbf{47.6} & \textbf{57.6} & \textbf{37.4} \\
32 & \textbf{24.3} & 46.6 & 56.4 & 37.2 \\
64 & 23.9 & 45.6 & 55.2 & 36.2 \\
\bottomrule
\end{tabular}
}
\vspace{-0.25em}
\textcolor{black}{\caption{\textbf{\method{} scales effectively up to 16 examples per \efa{}.} We train \llamabase{} on varying numbers of problem variants generated from each \efa{} and evaluate on MATH-500. Performance improves smoothly from 1 to 16 examples per \efa{}, with diminishing returns beyond that point. Bold numbers indicate the best performance for each metric.}
\label{tab:scaling_examples}}
\vspace{-1em}
\end{table}

\section{Ablation: Unit Tests Improve \efa{}-Based Data Augmentation Quality}
\label{sec:unit-tests-ablation}

Despite some errors in \efa{} generation, we find that the current \efa{}s are effectively improving performance. When we lower the quality by removing our unit tests, the performance gains from augmentation also decrease. As shown in \cref{tab:unit_tests_ablation}, applying unit tests consistently improves performance across all benchmarks and metrics. The unit tests provide an average improvement of 2.2 percentage points on MATH-500 Pass@1, 1.7 percentage points on FnEval November Pass@1, and 2.9 percentage points on FnEval December Pass@1.

In general, we believe there is a tradeoff between the level of noise in generated data and the cost of data generation, and \efa{}s occupy a generally useful point on the tradeoff curve. We can change the tradeoff and reduce noise by increasing the cost of filtering and data generation. These same issues occur with synthetic data generation approaches. The value of our approach is that data generation can be replaced with program execution rather than a call to a frontier LLM.

\begin{table}[t]
\centering
\resizebox{\linewidth}{!}{
\begin{tabular}{@{}l ccc ccc ccc@{}}
\toprule
& \multicolumn{3}{c}{MATH-500} & \multicolumn{3}{c}{FnEval (November)} & \multicolumn{3}{c}{FnEval (December)} \\
\cmidrule(l{0.5em}r{0.5em}){2-4} \cmidrule(l{0.5em}r{0.5em}){5-7} \cmidrule(l{0.5em}r{0.5em}){8-10}
Unit Tests & Pass @ 1 & Pass @ 10 & Maj @ 25 & Pass @ 1 & Pass @ 10 & Maj @ 25 & Pass @ 1 & Pass @ 10 & Maj @ 25 \\
\midrule
False & 20.4 & 55.2 & 35.4 & 24.1 & 54.7 & 35.7 & 22.0 & 55.2 & 37.1 \\
True & \textbf{22.6} & \textbf{58.0} & \textbf{37.8} & \textbf{24.9} & \textbf{56.6} & \textbf{38.3} & \textbf{25.5} & \textbf{57.2} & \textbf{40.0} \\
$\Delta$ & \plus{2.2} & \plus{2.8} & \plus{2.4} & \plus{0.8} & \plus{1.9} & \plus{2.6} & \plus{3.5} & \plus{2.0} & \plus{2.9} \\
\bottomrule
\end{tabular}
}
\vspace{-0.25em}
\caption{\textcolor{black}{\textbf{Unit tests improve \efa{}-based data augmentation quality.} We compare the performance of \efa{}-based data augmentation with and without the unit tests that filter out low-quality \efa{}s. The unit tests consistently improve performance across all benchmarks, demonstrating their effectiveness in maintaining data quality.}}
\label{tab:unit_tests_ablation}
\vspace{-1em}
\end{table} 

\section{Qualitative Examples}
\label{sec:qual-examples}
In this section, we display qualitative examples of \efa{}s across the MATH training set which were validated by our tests.
\begin{pabox}[box:algebra-example]{\efa{} (Algebra)}
\textbf{Original Problem}\\
Solve the equation:

$$
\sqrt{5 x-4}+\frac{15}{\sqrt{5 x-4}}=8
$$

\textbf{Original Solution}\\
Let $u=\sqrt{5 x-4}$. We can rewrite the equation in terms of $u$ :

$$
u+\frac{15}{u}=8
$$

Multiply through by $u$ to remove the fraction:

$$
u^2+15=8 u
$$

Reorganize into a standard quadratic form:

$$
u^2-8 u+15=0
$$

Factorize the quadratic:

$$
(u-5)(u-3)=0
$$

Thus, $u=5$ or $u=3$. For $u=5$ :

$$
\sqrt{5 x-4}=5 \Rightarrow 5 x-4=25 \Rightarrow 5 x=29 \Rightarrow x=\frac{29}{5}
$$

For $u=3$ :

$$
\sqrt{5 x-4}=3 \Rightarrow 5 x-4=9 \Rightarrow 5 x=13 \Rightarrow x=\frac{13}{5}
$$

Therefore, the solutions are $x=\frac{29}{5}, \frac{13}{5}$. Conclusion: The possible values of $x$ are $\frac{29}{5}$ and $\frac{13}{5}$, which satisfy the original equation when we substitute back, confirming the correctness of our solution.
\\
\\
\textbf{Inferred \efa{}}
\begin{lstlisting}
class Problem(BaseModel):
    coefficient: int
    constant: int
    term: int

    @classmethod
    def original(cls) ->Self:
        return cls(coefficient=5, constant=4, term=15)

    @classmethod
    def sample(cls) ->Self:
        coefficient = random.randint(3, 10)
        constant = random.randint(1, 5)
        term = random.randint(10, 20)
        return cls(coefficient=coefficient, constant=constant, term=term)

    def render(self) ->str:
        return f"""Solve the equation:
\\[ \\sqrt{{{self.coefficient}x-{self.constant}}} + \\frac{{{self.term}}}{{\\sqrt{{{self.coefficient}x-{self.constant}}}}} = 8. \\]"""

    def solve(self) ->str:
        u = sympy.symbols('u')
        equation = sympy.Eq(u + self.term / u, 8)
        u_values = sympy.solve(equation, u)
        x_solutions = []
        for u_val in u_values:
            x_equation = sympy.Eq(u_val ** 2, self.coefficient * sympy.
                symbols('x') - self.constant)
            x_solution = sympy.solve(x_equation, sympy.symbols('x'))
            x_solutions.extend(x_solution)
        return ', '.join(
            f'\\frac{{{sol.as_numer_denom()[0]}}}{{{sol.as_numer_denom()[1]}}}'
             for sol in x_solutions)
\end{lstlisting}
\textbf{Variant generated by \efa{}}\\
Solve the equation: $$\sqrt{7 x-5}+\frac{14}{\sqrt{7 x-5}}=8$$
\textit{Solution}:
$$\frac{23 - 8\sqrt2}{7}, \frac{8\sqrt2 + 23}{7}$$
\end{pabox}
\begin{pabox}[box:numbertheory-example]{\efa{} (Number Theory)}
\textbf{Original Problem} \\
How many positive divisors does $8!$ have? \\
\textbf{Original Solution} \\
First, calculate $8!: 8!=1 \times 2 \times 3 \times 4 \times 5 \times 6 \times 7 \times 8=40320$. Next, find the prime factorization of $40320: 40320=2^7 \times 3^2 \times 5^1 \times 7^1$. Now, apply the formula for counting the divisors: If $n=p^a \times q^b \times r^c \times \ldots$, then the number of divisors $t(n)$ is given by:

$$
t(n)=(a+1)(b+1)(c+1) \ldots
$$

Here $a=7, b=2, c=1, d=1$ for the primes $2,3,5$, and 7 respectively. Applying the formula:

$$
t(40320)=(7+1)(2+1)(1+1)(1+1)=8 \times 3 \times 2 \times 2=96
$$

Conclusion: The result is consistent with the factorial and prime factorization, providing a logically correct count of divisors.\\
\\
\textbf{Inferred \efa{}}
\begin{lstlisting}
class Problem(BaseModel):
    n: int

    @classmethod
    def original(cls) ->Self:
        return cls(n=8)

    @classmethod
    def sample(cls) ->Self:
        n = random.randint(4, 10)
        return cls(n=n)

    def render(self) ->str:
        return f'How many positive divisors does {self.n}! have?'

    def solve(self) ->str:
        factorial_value = math.factorial(self.n)
        factors = sympy.factorint(factorial_value)
        divisor_count = 1
        for exponent in factors.values():
            divisor_count *= exponent + 1
        return str(divisor_count)
\end{lstlisting}
\textbf{Variant generated by \efa{}}\\
How many positive divisors does $9!$ have? \\
\textit{Solution}: $$160$$
\end{pabox}
\begin{pabox}[box:probability-example]{\efa{} (Probability)}
\textbf{Original Problem} \\
Two 8-sided dice are tossed. What is the probability that the sum of the numbers shown on the dice is a prime number? Express your answer as a common fraction.
\\
\textbf{Original Solution} \\
Let $d_1$ and $d_2$ be the outcomes of the two 8-sided dice, where $d_1, d_2 \in \{1, 2, \dots, 8\}$.

The total number of possible outcomes in the sample space is:
\[
|\Omega| = 8 \times 8 = 64
\]
We want to find the number of outcomes where the sum $S = d_1 + d_2$ is a prime number. The smallest possible sum is $1+1=2$ and the largest is $8+8=16$. The prime numbers in this range are $2, 3, 5, 7, 11, \text{ and } 13$.

We enumerate the pairs $(d_1, d_2)$ corresponding to each prime sum:
\begin{itemize}
    \item[\textbf{Sum = 2:}] $\{(1,1)\}$ \hfill (1 way)
    \item[\textbf{Sum = 3:}] $\{(1,2), (2,1)\}$ \hfill (2 ways)
    \item[\textbf{Sum = 5:}] $\{(1,4), (4,1), (2,3), (3,2)\}$ \hfill (4 ways)
    \item[\textbf{Sum = 7:}] $\{(1,6), (6,1), (2,5), (5,2), (3,4), (4,3)\}$ \hfill (6 ways)
    \item[\textbf{Sum = 11:}] $\{(3,8), (8,3), (4,7), (7,4), (5,6), (6,5)\}$ \hfill (6 ways)
    \item[\textbf{Sum = 13:}] $\{(5,8), (8,5), (6,7), (7,6)\}$ \hfill (4 ways)
\end{itemize}
The total number of favorable outcomes is the sum of the ways for each prime:
\[
N_{\text{favorable}} = 1 + 2 + 4 + 6 + 6 + 4 = 23
\]
The probability is the ratio of the number of favorable outcomes to the total number of outcomes:
\[
P(\text{Sum is prime}) = \frac{N_{\text{favorable}}}{|\Omega|} = \frac{23}{64}
\]
The final answer is:
\[
\boxed{\frac{23}{64}}
\]
\\
\textbf{Inferred EFA}
\begin{lstlisting}
class Problem(BaseModel):
    sides: int

    @classmethod
    def original(cls) ->Self:
        return cls(sides=8)

    @classmethod
    def sample(cls) ->Self:
        sides = random.randint(4, 20)
        return cls(sides=sides)

    def solve(self) ->str:
        total_outcomes = self.sides * self.sides

        def is_prime(n: int) ->bool:
            if n <= 1:
                return False
            for i in range(2, int(math.sqrt(n)) + 1):
                if n % i == 0:
                    return False
            return True
        primal_sum_occurrences = 0
        for die1 in range(1, self.sides + 1):
            for die2 in range(1, self.sides + 1):
                sum_of_dice = die1 + die2
                if is_prime(sum_of_dice):
                    primal_sum_occurrences += 1
        probability = primal_sum_occurrences / total_outcomes
        fraction = sympy.Rational(primal_sum_occurrences, total_outcomes)
        return f'\\frac{{{fraction.numerator}}}{{{fraction.denominator}}}'

    def render(self) ->str:
        return (
            f'Two {self.sides}-sided dice are tossed. What is the probability that the sum of the numbers shown on the dice is a prime number? Express your answer as a common fraction.'
            )
\end{lstlisting}
\textbf{Variant generated by \method{}}
Two 19-sided dice are tossed. What is the probability that the sum of the numbers shown on the dice is a prime number? Express your answer as a common fraction. \\
\textit{Solution}:
$$\frac{105}{361}$$
\end{pabox}
\section{Experimental Details}
\subsection{Generating \efa{}s}
When generating \efa{}s, we use the prompt in \cref{efagen-prompt}.
To sample multiple candidates for \efa{}s, we use beam search with a temperature of 0.7 and a max generation length of 4096.
We extract the resulting \efa{}s from the LLMs response by looking for a markdown code block and extracting all markdown code blocks that have the necessary class structure.
\begin{pabox}[efagen-prompt]{Prompt for Inferring EFAs}
\begin{lstlisting}
# Instructions for Math Problem Functionalization

Your task is to convert a mathematical problem and its solution into a reusable Python class that can generate similar problems. Follow these steps:

1. Create a Python class that inherits from BaseModel with parameters that can vary in the problem. These parameters should capture the core numerical or mathematical values that could be changed while maintaining the same problem structure.

2. Implement the following required methods:
   - `original()`: A class method that returns the original problem's parameters
   - `sample()`: A class method that generates valid random parameters for a similar problem
   - `render()`: An instance method that produces the problem statement as a formatted string
   - `solve()`: An instance method that computes and returns the solution

3. For the `sample()` method:
   - Generate random parameters that maintain the problem's mathematical validity
   - Include appropriate constraints and relationships between parameters
   - Use reasonable ranges for the random values

4. For the `render()` method:
   - Format the problem statement using f-strings
   - Include proper mathematical notation using LaTeX syntax where appropriate
   - Maintain the same structure as the original problem

5. For the `solve()` method:
   - Implement the solution logic using the instance parameters
   - Return the final answer in the expected format (string, typically)
   - Include any necessary helper functions within the method

6. Consider edge cases and validity:
   - Ensure generated problems are mathematically sound
   - Handle special cases appropriately
   - Maintain reasonable complexity in generated problems

7. Do not import any libraries! The following libraries have been imported. Use fully qualified names for all imports:
    - pydantic.BaseModel is imported as `BaseModel`
    - random is imported as `random`
    - math is imported as `math`
    - numpy is imported as `np`
    - sympy is imported as `sympy` 
    - typing.Self is imported as `Self`

Example usage:
```python
problem = MyMathProblem.original()  # Get original problem
variant = MyMathProblem.sample()    # Generate new variant
question = variant.render()         # Get problem statement
answer = variant.solve()           # Compute solution
```

The goal is to create a class that can both reproduce the original problem and generate mathematically valid variations of the same problem type. 

# Example 1
## Problem Statement 
Evaluate $i^5+i^{-25}+i^{45}$.

## Solution
We have $i^5 = i^4\\cdot i = 1\\cdot (i) = i$.  We also have $i^{-25} = 1/i^{25} = 1/(i^{24}\\cdot i) = 1/[1\\cdot (i)] = 1/i = \\frac1{i}\\cdot\\frac{i}{i} = i/(-1) = -i$ and $i^{45} =
(i^{44})\\cdot i= 1\\cdot i =i$, and . So, adding these three results gives $i^5 + i^{-25} + i^{45} = i+-i+i = \\boxed{i}$.\nFinal Answer: The final answer is $\\boxed{ i }$.

## Functionalization
```python
class Problem(BaseModel):
    exponent1: int
    exponent2: int
    exponent3: int

    @classmethod
    def original(cls) -> Self:
        return cls(exponent1=5, exponent2=-25, exponent3=45)

    @classmethod
    def sample(cls) -> Self:
        exponent1 = random.randint(-100, 100)
        exponent2 = random.randint(-100, 100)
        exponent3 = random.randint(-100, 100)
        return cls(exponent1=exponent1, exponent2=exponent2, exponent3=exponent3)

    def render(self) -> str:
        return f"Evaluate $i^{{{self.exponent1}}} + i^{{{self.exponent2}}} + i^{{{self.exponent3}}}$."

    def solve(self) -> str:
        # Compute the values of i^n mod 4 cycle
        def compute_i_power(exp: int) -> complex:
            cycle = [1, 1j, -1, -1j]  # 1, i, -1, -i
            return cycle[exp % 4]

        # Compute each term
        term1 = compute_i_power(self.exponent1)
        term2 = compute_i_power(self.exponent2)
        term3 = compute_i_power(self.exponent3)

        # Calculate the sum
        result = term1 + term2 + term3

        # Express as LaTeX
        result_latex = (
            f"{result:.0f}" if result.imag == 0 else str(result).replace("j", "i")
        )
        return f"{result_latex}"
```

# Example 2
## Problem Statement
Altitudes $\overline{AX}$ and $\overline{BY}$ of acute triangle $ABC$ intersect at $H$.  If $\angle BAC = 43^\circ$ and $\angle ABC = 67^\circ$, then what is $\angle HCA$?
## Solution
First, we build a diagram:


size(150); defaultpen(linewidth(0.8));
pair B = (0,0), C = (3,0), A = (1.2,2), P = foot(A,B,C), Q = foot(B,A,C),H = intersectionpoint(B--Q,A--P);
draw(A--B--C--cycle);
draw(A--P^^B--Q);
pair Z;
Z = foot(C,A,B);
draw(C--Z);
label("$A$",A,N); label("$B$",B,W); label("$C$",C,E); label("$X$",P,S); label("$Y$",Q,E); label("$H$",H+(0,-0.17),SW);
label("$Z$",Z,NW);
draw(rightanglemark(B,Z,H,3.5));
draw(rightanglemark(C,P,H,3.5));
draw(rightanglemark(H,Q,C,3.5));


Since altitudes $\overline{AX}$ and $\overline{BY}$ intersect at $H$, point $H$ is the orthocenter of $\triangle ABC$.  Therefore, the line through $C$ and $H$ is perpendicular to
side $\overline{AB}$, as shown.  Therefore, we have $\angle HCA = \angle ZCA = 90^\circ - 43^\circ = \boxed{47^\circ}$.

## Functionalization
```python
class Problem(BaseModel):
    angle_BAC: int  # angle BAC in degrees
    angle_ABC: int  # angle ABC in degrees

    @classmethod
    def original(cls) -> Self:
        return cls(angle_BAC=43, angle_ABC=67)

    @classmethod
    def sample(cls) -> Self:
        # Generate random acute angles that form a valid triangle
        # Sum of angles must be less than 180
        angle1 = random.randint(30, 75)  # Keep angles acute
        angle2 = random.randint(30, 75)
        # Ensure the third angle is also acute
        if angle1 + angle2 >= 150:
            angle1 = min(angle1, 60)
            angle2 = min(angle2, 60)
        return cls(angle_BAC=angle1, angle_ABC=angle2)

    def solve(self) -> str:
        # The angle HCA is complementary to angle BAC
        # This is because H is the orthocenter and CH is perpendicular to AB
        angle_HCA = 90 - self.angle_BAC
        return f"{angle_HCA}"

    def render(self) -> str:
        return (
            f"Altitudes $\\overline{{AX}}$ and $\\overline{{BY}}$ of acute triangle $ABC$ "
            f"intersect at $H$. If $\\angle BAC = {self.angle_BAC}^\\circ$ and "
            f"$\\angle ABC = {self.angle_ABC}^\\circ$, then what is $\\angle HCA$?"
        )
```

# Example 3
## Problem Statement
On a true-false test of 100 items, every question that is a multiple of 4 is true, and all others are false. If a student marks every item that is a multiple of 3 false and all others true, how many of the 100 items will be correctly answered?
## Solution
The student will answer a question correctly if

Case 1: both the student and the answer key say it is true. This happens when the answer is NOT a multiple of 3 but IS a multiple of 4.

Case 2. both the student and the answer key say it is false. This happens when the answer IS a multiple of 3 but is NOT a multiple of 4.

Since the LCM of 3 and 4 is 12, the divisibility of numbers (in our case, correctness of answers) will repeat in cycles of 12. In the first 12 integers, $4$ and $8$ satisfy Case 1
and $3,6,$ and $9$ satisfy Case 2, so for every group of 12, the student will get 5 right answers. Since there are 8 full groups of 12 in 100, the student will answer at least $8
\cdot 5 = 40$ questions correctly. However, remember that we must also consider the leftover numbers 97, 98, 99, 100 and out of these, $99$ and $100$ satisfy one of the cases. So
our final number of correct answers is $40 + 2 = \boxed{42}$.

## Functionalization
```python
class Problem(BaseModel):
    total_questions: int  # Total number of questions
    multiple1: int  # First multiple (4 in original problem)
    multiple2: int  # Second multiple (3 in original problem)

    @classmethod
    def original(cls) -> Self:
        return cls(total_questions=100, multiple1=4, multiple2=3)

    @classmethod
    def sample(cls) -> Self:
        # Generate reasonable random parameters
        total = random.randint(50, 200)  # Reasonable test length
        # Choose coprimes or numbers with small LCM for interesting results
        mult1 = random.randint(2, 6)
        mult2 = random.randint(2, 6)
        while mult1 == mult2:  # Ensure different numbers
            mult2 = random.randint(2, 6)
        return cls(total_questions=total, multiple1=mult1, multiple2=mult2)

    def solve(self) -> str:
        def lcm(a: int, b: int) -> int:
            def gcd(x: int, y: int) -> int:
                while y:
                    x, y = y, x % y
                return x

            return abs(a * b) // gcd(a, b)

        # Find cycle length (LCM)
        cycle_length = lcm(self.multiple1, self.multiple2)

        # Count correct answers in one cycle
        correct_per_cycle = 0
        for i in range(1, cycle_length + 1):
            answer_key_true = i % self.multiple1 == 0
            student_true = i % self.multiple2 != 0
            if answer_key_true == student_true:
                correct_per_cycle += 1

        # Calculate complete cycles and remainder
        complete_cycles = self.total_questions // cycle_length
        remainder = self.total_questions % cycle_length

        # Calculate total correct answers
        total_correct = complete_cycles * correct_per_cycle

        # Add correct answers from remainder
        for i in range(1, remainder + 1):
            answer_key_true = i % self.multiple1 == 0
            student_true = i % self.multiple2 != 0
            if answer_key_true == student_true:
                total_correct += 1

        return str(total_correct)

    def render(self) -> str:
        return (
            f"On a true-false test of {self.total_questions} items, "
            f"every question that is a multiple of {self.multiple1} is true, "
            f"and all others are false. If a student marks every item that is "
            f"a multiple of {self.multiple2} false and all others true, how "
            f"many of the {self.total_questions} items will be correctly answered?"
        )
```

# Your Turn
Functionalize the following problem:

## Problem Statement
[% problem_statement %]

## Solution
[% solution %]

## Functionalization
\end{lstlisting}
\label{fig:efagen-prompt}
\end{pabox}
\subsection{\method{} Training Details}
\label{sec:training-details}
When doing rejection finetuning, we sample 20 candidate \efa{}s programs from the LLM for each seed problem during the rejection sampling phase.
We sample 20 variants from each \efa{}~in order to run the \texttt{\textbf{has\_dof(EFA)}} and \textbf{\texttt{is\_single\_valued(EFA)}} tests.
When finetuning on the \efa{}s that pass all tests, we use the the same prompt \cref{efagen-prompt} as the instruction and the extracted code of the \efa{}~as the response.
We use Transformers~\citep{transformers} and Llama-Factory~\citep{llamafactory} libraries for training.
We format all data in the Alpaca format~\citep{alpaca} as instruction-response pairs.
We use the Adam optimizer with a batch size of 16 and a cosine learning rate scheduler with a warmup ratio of 0.1 and train for 3 epochs in the FP16 datatype.
We apply LoRA to all linear layers with a rank of 16 and an alpha of 32, no bias, and a dropout of 0.05. 
We truncate all training examples to a maximum length of 4096 tokens with a batch size of 32.
\subsection{Math Inference Settings}
When doing 0-shot inference with \llama, we use the official Llama3.1 prompt in \cref{box:official-llama-3-0shot-prompt}.
When doing few-shot inference with \llama, we use a modified version of the official prompt, shown in \cref{box:unofficial-llama-3-nshot-prompt}.
When sampling multiple responses, we use beam search with a temperature of 0.7 and a max generation length of 2048.
When sampling a single response, we use beam search with a temperature of 0.0 and a max generation length of 2048.
In all cases, we check for equality of answers using the \href{https://github.com/huggingface/Math-Verify}{math-verify} library.
\begin{pabox}[box:official-llama-3-0shot-prompt]{Llama3.1 0-shot MATH Prompt}
\begin{lstlisting}
Solve the following math problem efficiently and clearly:

- For simple problems (2 steps or fewer):
Provide a concise solution with minimal explanation.

- For complex problems (3 steps or more):
Use this step-by-step format:

## Step 1: [Concise description]
[Brief explanation and calculations]

## Step 2: [Concise description]
[Brief explanation and calculations]

...

Regardless of the approach, always conclude with:

Therefore, the final answer is: $\boxed{answer}$. I hope it is correct.

Where [answer] is just the final number or expression that solves the problem.

Problem: {{ instruction }}
\end{lstlisting}
\end{pabox}

\begin{pabox}[box:unofficial-llama-3-nshot-prompt]{Llama3.1 N-shot MATH Prompt}
\begin{lstlisting}
Solve the following math problem efficiently and clearly:

- For simple problems (2 steps or fewer):
Provide a concise solution with minimal explanation.

- For complex problems (3 steps or more):
Use this step-by-step format:

\#\# Step 1: [Concise description]
[Brief explanation and calculations]

\#\# Step 2: [Concise description]
[Brief explanation and calculations]

...

Regardless of the approach, always conclude with:

Therefore, the final answer is: $\boxed{answer}$. I hope it is correct.

Where [answer] is just the final number or expression that solves the problem.

Here are some examples:
{% for few_shot_example in few_shot_examples %}
Problem: {{ few_shot_example.instruction }}
{{ few_shot_example.response }}
{% endfor %}

Problem: {{ instruction }}
\end{lstlisting}
\end{pabox}
\subsection{Math Training Details}
\label{sec:math-training}
We use the same hyperparameters and chat data format as in \cref{sec:training-details}, except we cutoff training data over 2048 tokens.
However, we use a simpler prompt template, shown in \cref{box:minimal-prompt} to format the teacher responses.
When annotating with a \llama~teacher, we sample 5 responses per math problem with a temperature of 0.7.
We check for equality of answers using the \href{https://github.com/huggingface/Math-Verify}{math-verify} library.
\begin{pabox}[box:minimal-prompt]{Minimal instruction-tuning prompt used for augmentation experiments}
\begin{lstlisting}
Question: {{ question }}
Step-by-step Answer
\end{lstlisting}
\end{pabox}

\end{document}